\documentclass{article}

\usepackage[utf8]{inputenc} 
\usepackage[T1]{fontenc}    
\usepackage{hyperref}       
\usepackage{url}            
\usepackage{booktabs}       
\usepackage{amsfonts}       
\usepackage{nicefrac}       
\usepackage{microtype}      
\usepackage{xcolor}         
\usepackage{graphicx}
\usepackage{multirow}
\usepackage{multicol}
\usepackage{wrapfig}
\usepackage{enumitem}
\usepackage{amsmath}
\usepackage{float}
\usepackage{cite}
\setlist[itemize]{noitemsep, topsep=0pt}
\usepackage{pifont}

\title{MAMAT: 3D Mamba-Based Atmospheric Turbulence Removal and its Object Detection Capability}

\author{\small Paul Hill, Zhiming Liu and Nantheera Anantrasirichai\\
\small Visual Information Laboratory\\
\small University of Bristol
}
\date{}

\begin{document}

\maketitle

\begin{abstract}
  Restoration and enhancement are essential for improving the quality of videos captured under atmospheric turbulence conditions, aiding visualization, object detection, classification, and tracking in surveillance systems. In this paper, we introduce a novel Mamba-based method, the 3D Mamba-Based Atmospheric Turbulence Removal (MAMAT), which employs a dual-module strategy to mitigate these distortions. The first module utilizes deformable 3D convolutions for non-rigid registration to minimize spatial shifts, while the second module enhances contrast and detail. Leveraging the advanced capabilities of the 3D Mamba architecture, experimental results demonstrate that MAMAT outperforms state-of-the-art learning-based methods, achieving up to a 3\% improvement in visual quality and a 15\% boost in object detection. It not only enhances visualization but also significantly improves object detection accuracy, bridging the gap between visual restoration and the effectiveness of surveillance applications. 
\end{abstract}

\section{Introduction}

Atmospheric turbulence distortions, caused by significant temperature gradients between the ground and air, can severely degrade image and video quality. This phenomenon leads to rapidly rising air layers, altering the index of refraction along the optical path and resulting in visual distortions such as blurriness, ripples, and intensity fluctuations. These effects pose significant challenges for long-range surveillance, particularly in identifying distant objects.

Modeling atmospheric turbulence in video footage is a computationally demanding and complex task, primarily because it is an ill-posed problem with distortions that vary across space and time. Model-based approaches often hinder real-time implementation, limiting their effectiveness in dynamic settings \cite{6178259,Anantrasirichai:Atmospheric:2013, anantrasirichai2018atmospheric, CHEN2020106131}. Additionally, merging multiple images to enhance quality can introduce artifacts, especially in the presence of moving objects, due to misalignment. Consequently, there is growing interest in leveraging deep learning-based video restoration techniques, which offer more robust solutions.

The initial deep learning advancements in atmospheric turbulence correction predominantly relied on convolutional neural networks (CNNs) \cite{Gao:Atmospheric:2019,photonics10060666,anantrasirichai2023atmospheric,Chak:Subsampled:2021,Wang:deep:2021,Yasarla:ATNet:2021}. More recently, there has been a shift towards Transformer-based models, which have demonstrated improved performance in turbulence mitigation \cite{zhang2024spatio,wang2024real,wu2024semi,Zhang_TMT,Saha:Turb:2024,zou2024deturb}. Given the complexity of atmospheric distortions, modern systems typically incorporate dual modules: one dedicated to frame alignment and enhancement \cite{anantrasirichai2023atmospheric, zou2024deturb} and another focused on correcting tilt and blur \cite{Yasarla:ATNet:2021,Zhang_TMT}. These modules collectively demonstrate significant efficacy in addressing the intricate problem of atmospheric turbulence, highlighting the potential of integrated approaches for improving visual clarity and stability.

Though Transformer-based methods show promising results, the computational demands present challenges in processing complex, distorted images. Structured State Space sequence (S4) models offer efficient alternatives, excelling in long-sequence data analysis. The Mamba framework~\cite{Zhu:visionmamba:2024} enhances S4, particularly in handling time-variant operations, and has shown promising results in image analysis and computer vision. To date, Mamba has not been adapted for atmospheric turbulence removal, highlighting a promising direction for future research.

In this paper, we propose a novel Mamba-based method for atmospheric turbulence mitigation, referred to as 3D \textbf{Mam}ba-Based \textbf{A}tmospheric \textbf{T}urbulence Removal (MAMAT). We employ a dual-module strategy: the first module performs non-rigid registration to minimize spatial shifts between frames caused by random temporal permutations, utilizing deformable 3D convolutions. The second module enhances contrast, sharpness, and textural details using the 3D Mamba architecture~\cite{gong2024nnmamba}, known for its efficiency in processing spatiotemporal data. By utilizing this framework, MAMAT effectively addresses complex distortions in video sequences, setting a new benchmark for clarity and precision in turbulence mitigation.

It is important to note that existing methods may effectively mitigate atmospheric turbulence but may not optimize imagery for crucial automatic object detection in surveillance. Techniques that reduce visual distortions, such as wavy or ripple effects, could inadvertently remove vital features necessary for object identification. Therefore, we demonstrate that our proposed MAMAT not only enhances visual quality but also significantly improves object detection accuracy, effectively bridging the gap between visual restoration and practical surveillance applications.

In summary, our main contributions can be listed as follows:
\begin{itemize}[noitemsep,topsep=0pt,leftmargin=10pt, itemindent=0pt]
\item We propose a novel framework, MAMAT, for restoring long-range videos affected by spatiotemporal distortions.
\item We apply deformable 3D convolutions in all feature scales, providing flexibility in capturing the different shapes of distorted objects.
\item MAMAT is the first method to leverage the 3D Mamba block for atmospheric turbulence removal.
\item We demonstrate that MAMAT enhances visual quality and also improves object detection accuracy, both of which are crucial for various applications.
\end{itemize}

\section{Related Work}

\paragraph{Atmospheric Turbulence Mitigation.}
Early work in deep learning-based approaches employed convolutional neural networks, including the Gaussian denoiser architecture \cite{Gao:Atmospheric:2019}, the UNet architecture \cite{Mao:accelaring:2021}, and stacked hourglass networks~\cite{photonics10060666,anantrasirichai2023atmospheric}. A WGAN (Wasserstein generative adversarial network) is employed in \cite{Chak:Subsampled:2021}, where the multiple lucky frames are fed into the UNet generator. 
DATUM \cite{zhang2024spatio} introduces deformable attention alignment and temporal fusion via multi-head temporal channel self-attention.
Turb-Seg-Res \cite{Saha:Turb:2024} stabilizes dynamic video frames using normalized cross-correlation, segments moving objects, and refines the background via adaptive filtering before merging it with a separately extracted foreground. DeTurb \cite{zou2024deturb} employs a pyramid architecture with deformable 3D convolutions to eliminate spatial-temporal distortions. It then reconstructs a restored image using a multi-scale architecture composed of 3D Swin Transformers, thereby improving contrast and reducing blur. For a more comprehensive review of recent deep learning techniques for atmospheric turbulence removal, see~\cite{Hill2025}.

\vspace{-3mm}
\paragraph{Object Detection.}
In clear mediums, well-known CNN-based methods such as Faster R-CNN \cite{Ren:FasterRCNN:2017} and RetinaNet \cite{Lin:Focal:2017} utilize multi-scale feature extraction. YOLO (You Only Look Once), particularly its latest iteration, YOLOv11 \cite{khanam2024yolov11}, is the most popular real-time detector, incorporating advanced techniques to enhance performance.

For atmospheric turbulence conditions, traditional object detection methods primarily focus on long-distance objects that often appear small and lack detail \cite{Gilles:Detection:2018, Zhang:Stabilization:2018}. These methods are limited by the reduced clarity and specificity of the images. Despite these challenges, deep learning techniques specifically designed for these conditions are rare. Face recognition using Generative Adversarial Networks (GANs) \cite{Lau:ATFaceGAN:2020} has achieved some success, although this method struggles in complex scenarios involving various object types. To extend the capabilities, Hu et al. \cite{Hu:object:2023} have improved Faster R-CNN and YOLOv4 \cite{Bochkovskiy:YOLOv4:2020} with deformable convolutional layers to address frame displacement. Uzun et al. \cite{Uzun:Augmentation:2022} demonstrated that YOLOR \cite{Wang:YOLOR:2021} outperforms VfNet \cite{Zhang:VarifocalNet:2021} and TOOD \cite{Feng:TOOD:2021} in both precision and speed, showing a promising direction for enhancing object detection in turbulent conditions.

\section{Methodology}
\label{sec:method}

\begin{figure*}[t]
\begin{center}
   \includegraphics[width=\linewidth]{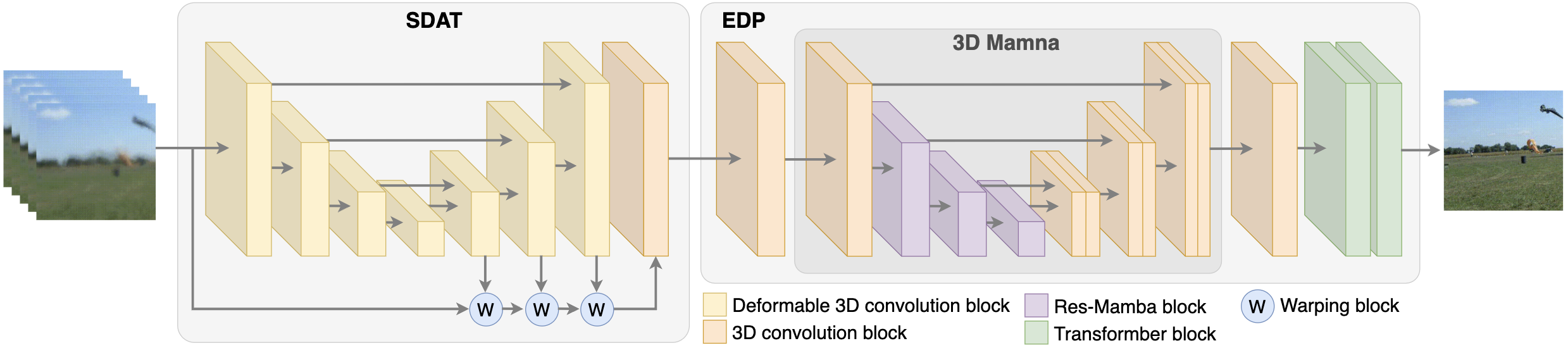}
\end{center}
   \caption{Diagram of the proposed MAMAT framework}
\label{fig:diagram}
\end{figure*}

The diagram of the proposed MAMAT framework is shown in Fig.~\ref{fig:diagram}, comprising two modules: (i) the \textbf{S}patial \textbf{D}isplacement \textbf{A}cross \textbf{T}ime Removal (SDAT) module, which removes spatial displacements that occur over time, and (ii) the \textbf{E}nhancement and \textbf{D}etail \textbf{P}reserve (EDP) module, which enhances visual quality spatially and temporally with 3D Mamba.

\subsection{SDAT: Spatial Displacement Across Time Removal Module}
\label{ssec:SDAT}
Addressing the challenge of atmospheric turbulence, particularly the wavy effect where features shift from their true positions across frames, requires robust solutions. Inspired by Zhang et al.~\cite{Zhang_TMT}, which utilizes a UNet architecture with depth-wise 3D convolutions to estimate motion across frames, we propose an enhancement by substituting these with deformable 3D convolutions at all scales. Deformable 3D convolutions offer the necessary flexibility to accurately capture the distorted shapes of objects across various spatial and temporal scales within the UNet framework. This adaptation enables more effective extraction of pertinent features from scenes affected by turbulence.

\paragraph{Deformable 3D convolution.} 
Denoting $p_0$ as a point on the output feature map $y$ and $p_n$ as the $n$-th location on the convolution grid $\mathcal{G}$, which has a size of 3$\times$3$\times$3, of the input $x$, the deformable 3D convolution is defined as follows:
\begin{equation}
y(p_0) = \sum_{p_n \in \mathcal{G}} w(p_n) \cdot x(p_0 + p_n + \Delta p_n),
\label{DConv3D}
\end{equation}
where $w$ denotes the convolution weights. $\Delta p_n$ represents learnable offsets, allowing the convolutional kernel to adapt to shape variations in the input, thereby optimizing feature extraction of dynamic contents.

\paragraph{Architecture.} 
Our SDAT module employs a multi-scale UNet-like architecture with four depth levels, using deformable 3D convolutions combined with ReLU activation functions. Following the strategies from \cite{Zhang_TMT}, the encoder utilizes progressively smaller kernel sizes to optimize feature detection across scales: starting with a size of 7 at the first level and reducing to sizes 5 and 3 in subsequent levels, allowing the network to effectively process a range of distortion patterns from broad to localized disturbances. The decoder employs deformable 3D transposed convolutions with a consistent kernel size of 3 across all levels.

\subsection{EDP: Enhancement and Detail Preservation Module}

The core implementation of this module is the 3D Mamba block, which our study found to outperform the 3D Swin Transformer~\cite{liu2021swin}, echoing findings in \cite{Hill2025}. Mamba is based on the Structured State Space sequence (S4) model, which adapts State Space Model (SSM) parameters based on the input and resolves the memory and gradient issues typically associated with SSM.

\paragraph{Architecture.} 
We first use a 3D convolution block to adjust the number of channels as required by the 3D Mamba input. For 3D Mamba, a UNet-like architecture with three depth levels is implemented to capture long-range dependencies between the encoder and decoder. Inspired by nnMamba~\cite{gong2024nnmamba}, the encoder consists of three residual Mamba blocks (Res-Mamba), while the decoder includes three double convolution blocks. The Res-Mamba block combines double convolution blocks, skip connections, and a Mamba-in-convolution block. The latter incorporates SSM between convolutional layers, as shown in Eq.~\ref{eq:ssminconv}.
\begin{equation}
F_{\text{o}} = \text{Conv}_{1}(\text{SSM}(\text{Conv}_{1}(F_{\text{i}}))) + \text{Conv}_{1}(F_{\text{i}}),
\label{eq:ssminconv}
\end{equation}
where $F_{\text{i}}$ and $F_{\text{o}}$ are the 3D input and output feature maps, respectively. $\text{Conv}_{1}(\cdot)$ denotes a convolutional layer with a kernel size of 1$\times$1$\times$1, followed by batch normalization and ReLU activation. SSM($\cdot$) is a selective SSM layer~\cite{Gu:mamba:2023} trained with four augmented inputs, each flipped in one of four different directions. Finally, the 3D convolution block and two Transformer blocks are applied for channel adjustment and final refinement.

\paragraph{Selective State Space Model.} 
SSM($\cdot$) maps an input $x_t \in \mathbb{R}$ to an output $y_t \in \mathbb{R}$ via an implicit latent state $h_t \in \mathbb{R}^d$, which can be expressed as follows:
\begin{equation}
h_t  =\overline{\mathbf{A}} h_{t-1}+\overline{\mathbf{B}} x_t, ~~~~ y_t  =\mathbf{C} h_t, ~~~ \mathbf{C} \in \mathbb{R}^{1 \times d}
\label{eq: discre_ssm}
\end{equation}
To integrate SSM into deep learning-based models, the discretized parameters $\overline{\mathbf{A}}$ and $\overline{\mathbf{B}}$ are formulated as:
\begin{equation}
\begin{aligned}
& \overline{\mathbf{A}}=\exp (\Delta \mathbf{A}) \\
& \overline{\mathbf{B}}=(\Delta \mathbf{A})^{-1}(\exp (\mathbf{A})-\mathbf{I}) \cdot \Delta \mathbf{B},
\end{aligned}
\end{equation}
where $\Delta \in \mathbb{R}$ is a timescale parameter. With \eqref{eq: discre_ssm}, the discretized SSM can be computed as a convolution.

\subsection{Loss Function}

Atmospheric turbulence can introduce outliers in pixel-wise loss calculations, necessitating the use of Charbonnier loss. This loss function, which effectively combines the advantages of both $\ell_1$ and $\ell_2$ losses, is particularly adept at handling outliers. It is defined as follows:
\begin{equation} 
L_\text{Char}(x, y) = \sqrt{(x-y)^2 + \epsilon^2}, 
\label{Loss} 
\end{equation}
where $x$ and $y$ are the predicted and true values, respectively. $\epsilon$ is a small constant (e.g., $1e-3$) added to ensure numerical stability. The Charbonnier loss provides a smooth gradient, even in the presence of small errors, making it highly suitable for dealing with the subtle yet significant discrepancies characteristic of turbulence-affected images. 


\begin{table}[t]
    \centering
    \caption{Comparison of average PSNR and SSIM for different methods. `Distorted' denotes the original raw videos. \textbf{Bold} indicates the best performance.}
    \begin{tabular}{ccc}
        \toprule
        {Method} & {Average PSNR} $\uparrow$ & {Average SSIM} $\uparrow$ \\
        \midrule
        Distorted & 30.42 & 0.4704 \\
        TMT~\cite{Zhang_TMT} & 30.83 & 0.6344 \\
        DATUM~\cite{zhang2024spatio} & 30.87 & 0.6347 \\
        MAMAT (ours) & \textbf{30.99} & \textbf{0.6514} \\
        \bottomrule
    \end{tabular} 
    \label{tab:psnr_ssim}
\end{table}

\section{Datasets and Experiment settings}
\label{sec:experiments}

\paragraph{Datasets.}
Our model was trained using both real and synthetic turbulence distortions. For real turbulence, we utilized the CLEAR dataset~\cite{Anantrasirichai:Atmospheric:2013}, which comprises 12 sequences exhibiting a range of atmospheric turbulence, from subtle to strong distortions. To ensure dataset balance, we limited the number of frames in each sequence to 200. Frame-by-frame pseudo ground truth was generated using the technique described in \cite{anantrasirichai2018atmospheric}.

For synthetic turbulence, we employed the COCO dataset \cite{lin2014microsoft} in conjunction with the Phase-to-Space (P2S) Transform method \cite{Mao:accelaring:2021} to create the distortion effects. This method transforms spatially varying convolutions into invariant ones using learned basis functions derived from known turbulence models. The P2S Transform then applies these functions to convert phase representations into spatial turbulence effects, effectively simulating atmospheric disturbances in imaging. We generated turbulent videos consisting of 50 frames with the same spatial dimensions.

For object detection testing, we used only the validation set of the COCO dataset, as it provides ground truth. It has 5,000 images with for 69 object classes (including background) and bounding boxes. We applied the synthetic turbulence distortions similar to above explanation.
We evaluate object detection performance using Average Precision (AP), calculated as the area under the Precision-Recall curve. This metric aggregates precision-recall values across different Intersection over Union (IoU) thresholds, ranging from 0.5 to 0.95 in increments of 0.05. We report the mean AP across all IoU thresholds as AP@[IoU=0.5:0.95].

\vspace{-3mm}
\paragraph{Experiment settings.}
We randomly cropped the input sequences to dimensions of  256$\times$256 during training. This approach ensured a broad and randomized spatial variation within the training dataset. We utilized the Adam optimizer, starting with a learning rate of 0.0001. The models were trained for 200 epochs with no early stopping. We employ a sliding window technique to process videos, which involves taking a group of frames surrounding the current frame as input and outputting a single frame. This process is applied consistently across the entire sequence during both training and testing phases. As discussed in \cite{anantrasirichai2023atmospheric}, using five neighboring frames provides the optimal balance, yielding the best quantitative PSNR results.

The proposed framework was developed using Python, leveraging PyTorch and CUDA for computational efficiency. All training and testing processes were executed on a high-performance computing system equipped with Nvidia P100 GPUs.

\begin{figure}[t]
\begin{center}
   \includegraphics[width=0.7\columnwidth]{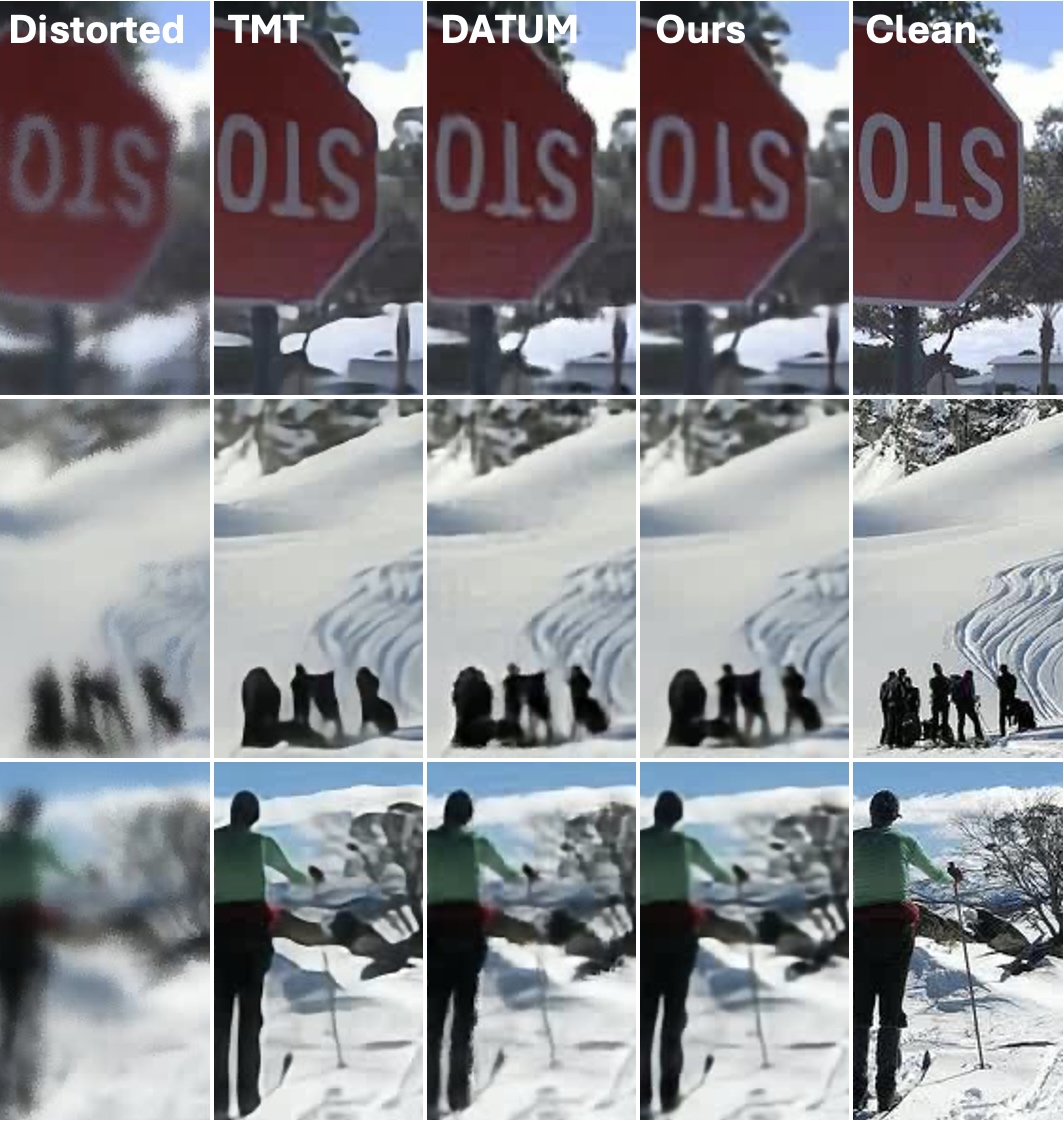}
\end{center}
   \caption{Subjective results comparing (from left to right) the distorted frame, the restored frames using TMT, DATUM, and our MAMAT, as well as the clean frame used as ground truth.}
\label{fig:results1}
\end{figure}

\section{Results and Discussion}
\label{sec:results} 

\subsection{Atmospheric turbulence removal}
\begin{figure}[t]
\begin{center}
   \includegraphics[width=0.7\columnwidth]{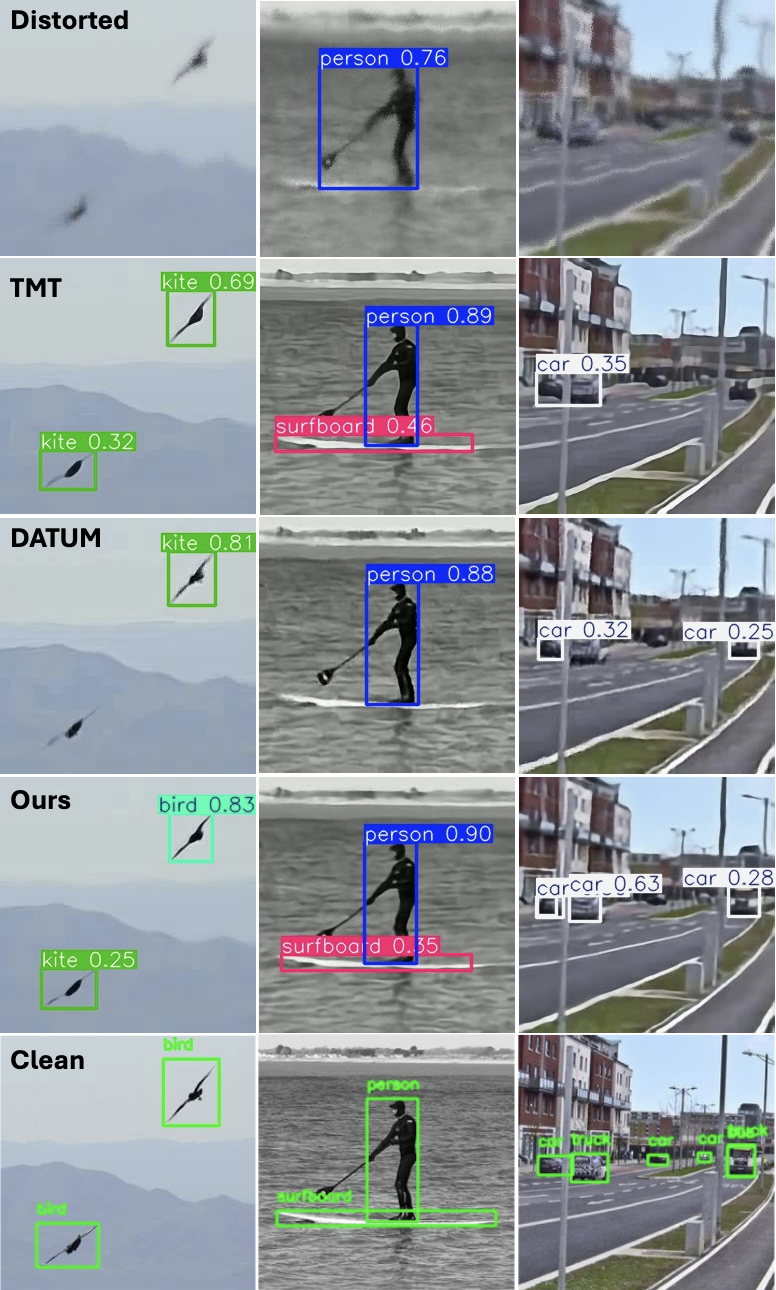}
\end{center}
   \caption{Subjective results of applying the YOLOv11 large model to the outputs of different restoration methods, as well as to distorted and clean videos (ground truth).}
\label{fig:results}
\end{figure}

We evaluated the performance of our proposed MAMAT against two leading deep learning-based methods for atmospheric turbulence removal: TMT~\cite{Zhang_TMT} and DATUM~\cite{zhang2024spatio}. The comparative results, including the average PSNR and SSIM values across all frames of the testing videos, are summarized in Table~\ref{tab:psnr_ssim}. The results indicate that MAMAT consistently surpasses both TMT and DATUM in terms of image quality and structural similarity. Fig.\ref{fig:results1} shows examples of the restored results, with additional visual comparisons from different methods available in Fig.\ref{fig:results}. TMT produces sharp edges but blurs textural details, while DATUM preserves finer details better. Our MAMAT balances both approaches and achieves better shapes of small objects.

We conducted an ablation study by replacing the 3D Mamba block with a 3D Swin Transformer \cite{gong2024nnmamba}, resulting in a decrease of 0.51\% in average PSNR and 0.97\% in SSIM.

\begin{table}[t]
    \centering
    \small
    \caption{Performance comparison across different detectors. `Distorted' denotes the original raw videos. `Clean' denotes the clean videos (ground truth), indicating the upper bound of detection performance. \textbf{Bold} indicates the best performance.} \vspace{5mm}
    \begin{tabular}{lccc}
    \toprule
    \multirow{2}{*}{Detector} & \multirow{2}{*}{Restored Method} & \multicolumn{2}{c}{AP@[IoU=0.5:0.95]} \\ \cline{3-4}
    & & all objects & small objects \\
    \midrule 
    \multirow{5}{*}{\parbox{2.7cm}{Faster R-CNN \\ \cite{Ren:FasterRCNN:2017}}}& Distorted & 0.043 & 0.007 \\
    & TMT~\cite{Zhang_TMT} & 0.197 & 0.053 \\
    & DATUM~\cite{zhang2024spatio} & 0.217 & 0.068 \\
    & MAMAT & \textbf{0.227} & \textbf{0.075} \\
    & {\color{gray}Clean} & {\color{gray}{0.433}} & {\color{gray}0.253} \\
    \hline
    \multirow{5}{*}{RetinaNet \cite{Lin:Focal:2017}}& Distorted & 0.066 & 0.023 \\
    & TMT~\cite{Zhang_TMT} & 0.218 & 0.101 \\
    & DATUM~\cite{zhang2024spatio} & 0.247 & \textbf{0.113} \\
    & MAMAT & \textbf{0.249} & 0.110 \\
    & {\color{gray}Clean} & {\color{gray}{0.471}} & {\color{gray}0.351} \\
    \hline
    \multirow{5}{*}{\parbox{2.7cm}{YOLOv11 \cite{khanam2024yolov11} \\ Small model}} & Distorted & 0.137 & 0.282 \\
    & TMT~\cite{Zhang_TMT} & 0.334 & 0.497 \\
    & DATUM~\cite{zhang2024spatio} & 0.334 & 0.497 \\
    & MAMAT & \textbf{0.355} & \textbf{0.511} \\
    & {\color{gray}Clean} & {\color{gray}{0.462}} & {\color{gray}0.630} \\
    \hline
    \multirow{5}{*}{\parbox{2.7cm}{YOLOv11 \cite{khanam2024yolov11} \\ Large model}} & Distorted & 0.258 & 0.402 \\
    & TMT~\cite{Zhang_TMT} & 0.427 & 0.587 \\
    & DATUM~\cite{zhang2024spatio} & 0.422 & 0.593 \\
    & MAMAT & \textbf{0.429} & \textbf{0.617} \\
    & {\color{gray}Clean} & {\color{gray}{0.591}} & {\color{gray}0.731} \\
    \bottomrule
    \end{tabular}
    \label{tab:detector_performance}
\end{table}

\subsection{Object detection}

We employed four pretrained models to evaluate the quality of restored videos in terms of object detection improvement: Faster R-CNN \cite{Ren:FasterRCNN:2017}, RetinaNet \cite{Lin:Focal:2017}, and the extra-small and extra-large models of YOLOv11 \cite{khanam2024yolov11}. The pretrained Faster R-CNN and RetinaNet models are available via Detectron2\footnote{https://github.com/facebookresearch/detectron2}, while the pretrained YOLOv11 models can be found at Ultralytics\footnote{https://github.com/ultralytics/ultralytics}.


The results, expressed in mAP@[IoU=0.5:0.95], are presented in Table~\ref{tab:detector_performance} for all object sizes as well as for small objects specifically. All restoration models significantly improve object detection performance un atmospheric turbulence conditions, with our MAMAT model achieving the highest mAP values. Using the best detector, YOLOv11 extra-large model, the performance improves substantially, with average AP values increasing by 66\% compared to raw distorted videos. While MAMAT enhances the average AP from DATUM by 1.7\% for all objects, it achieves a 4\% improvement for small objects.

The subjective analysis presented in Fig.~\ref{fig:results} illustrates the performance improvement across different scenarios: i) an easy scenario where all models accurately detect the person; ii) a medium difficulty scenario where only certain methods correctly identify the surfboard and some cars; iii) a challenging scenario where solely our model successfully detects the bird in the top right corner and where all models fail to recognize distant cars. While TMT renders images with sharp edges, it tends to lose fine details, so many small objects are missed or detected incorrectly. Subjectively, the outputs from DATUM and our model appear similar, but our method distinctly excels in preserving essential features, thereby enhancing object detection and achieving superior results.


\section{Conclusions}
This paper presents a new learning-based method to mitigate atmospheric turbulence distortions, called MAMAT. The proposed method integrates two modules, both based on multiscale UNet-like architectures. The first module employs deformable 3D convolutions to register features between consecutive frames, while the second module leverages 3D Mamba to capture spatial-temporal dependencies, enhancing contrast and details in the videos. Experimental results show that MAMAT not only enhances visual quality but also significantly improves detection performance.

\clearpage
\bibliographystyle{unsrt}
\bibliography{main}

\begin{thebibliography}{10}

\bibitem{6178259}
Xiang Zhu and Peyman Milanfar.
\newblock Removing atmospheric turbulence via space-invariant deconvolution.
\newblock {\em PAMI}, 35(1), 2013.

\bibitem{Anantrasirichai:Atmospheric:2013}
N.~Anantrasirichai, A.~Achim, N.G. Kingsbury, and D.R. Bull.
\newblock Atmospheric turbulence mitigation using complex wavelet-based fusion.
\newblock {\em IEEE TIP}, 22(6):2398--2408, 2013.

\bibitem{anantrasirichai2018atmospheric}
Nantheera Anantrasirichai, Alin Achim, and David Bull.
\newblock Atmospheric turbulence mitigation for sequences with moving objects using recursive image fusion.
\newblock In {\em ICIP}, pages 2895--2899, 2018.

\bibitem{CHEN2020106131}
Gongping Chen, Zhisheng Gao, Qiaolu Wang, and Qingqing Luo.
\newblock Blind de-convolution of images degraded by atmospheric turbulence.
\newblock {\em Applied Soft Computing}, 89:106131, 2020.

\bibitem{Gao:Atmospheric:2019}
Jing Gao, N.~Anantrasirichai, and David Bull.
\newblock Atmospheric turbulence removal using convolutional neural network.
\newblock In {\em arXiv:1912.11350}, 2019.

\bibitem{photonics10060666}
Jiuming Cheng, Wenyue Zhu, Jianyu Li, Gang Xu, Xiaowei Chen, and Cao Yao.
\newblock Restoration of atmospheric turbulence-degraded short-exposure image based on convolution neural network.
\newblock {\em Photonics}, 10(6), 2023.

\bibitem{anantrasirichai2023atmospheric}
Nantheera Anantrasirichai.
\newblock Atmospheric turbulence removal with complex-valued convolutional neural network.
\newblock {\em Pattern Recognition Letters}, 171:69--75, 2023.

\bibitem{Chak:Subsampled:2021}
Wai~Ho Chak, Chun~Pong Lau, and Lok~Ming Lui.
\newblock Subsampled turbulence removal network.
\newblock {\em Mathematics, Computation and Geometry of Data}, 1(1):1--33, 2021.

\bibitem{Wang:deep:2021}
Kaiqiang Wang, MengMeng Zhang, Ju~Tang, Lingke Wang, et~al.
\newblock Deep learning wavefront sensing and aberration correction in atmospheric turbulence.
\newblock {\em PhotoniX}, 2(8), 2021.

\bibitem{Yasarla:ATNet:2021}
Rajeev Yasarla and Vishal~M. Patel.
\newblock Learning to restore images degraded by atmospheric turbulence using uncertainty.
\newblock In {\em ICIP}, pages 1694--1698, 2021.

\bibitem{zhang2024spatio}
Xingguang Zhang, Nicholas Chimitt, Yiheng Chi, Zhiyuan Mao, and Stanley~H Chan.
\newblock Spatio-temporal turbulence mitigation: A translational perspective.
\newblock In {\em CVPR}, 2024.

\bibitem{wang2024real}
Xijun Wang, Santiago L{\'o}pez-Tapia, and Aggelos~K Katsaggelos.
\newblock Real-world atmospheric turbulence correction via domain adaptation.
\newblock {\em arXiv preprint arXiv:2402.07371}, 2024.

\bibitem{wu2024semi}
Yubo Wu, Kuanhong Cheng, Ting Cao, Dong Zhao, and Junhuai Li.
\newblock Semi-supervised correction model for turbulence-distorted images.
\newblock {\em Optics Express}, 32(12):21160--21174, 2024.

\bibitem{Zhang_TMT}
Xingguang Zhang, Zhiyuan Mao, Nicholas Chimitt, and Stanley~H. Chan.
\newblock Imaging through the atmosphere using turbulence mitigation transformer.
\newblock {\em TCI}, 10:115--128, 2024.

\bibitem{Saha:Turb:2024}
Ripon~Kumar Saha, Dehao Qin, Nianyi Li, Jinwei Ye, and Suren Jayasuriya.
\newblock {Turb-Seg-Res: A} segment-then-restore pipeline for dynamic videos with atmospheric turbulence.
\newblock In {\em CVPR}, 2024.

\bibitem{zou2024deturb}
Z.~Zou and N.~Anantrasirichai.
\newblock Deturb: Atmospheric turbulence mitigation with deformable 3d convolutions and 3d swin transformers.
\newblock In {\em Proceedings of the Asian Conference on Computer Vision}, 2024.

\bibitem{Zhu:visionmamba:2024}
Lianghui Zhu, Bencheng Liao, Qian Zhang, Xinlong Wang, et~al.
\newblock Vision mamba: efficient visual representation learning with bidirectional state space model.
\newblock In {\em ICML}, 2024.

\bibitem{gong2024nnmamba}
Haifan Gong, Luoyao Kang, Yitao Wang, Xiang Wan, and Haofeng Li.
\newblock {nnMamba: 3D} biomedical image segmentation, classification and landmark detection with state space model.
\newblock In {\em ISBI}, 2025.

\bibitem{Mao:accelaring:2021}
Zhiyuan Mao, Nicholas Chimitt, and Stanley~H. Chan.
\newblock Accelerating atmospheric turbulence simulation via learned phase-to-space transform.
\newblock In {\em ICCV}, 2021.

\bibitem{Hill2025}
P.~Hill, N.~Anantrasirichai, A.~Achim, et~al.
\newblock Deep learning techniques for atmospheric turbulence removal: a review.
\newblock {\em Artificial Intelligence Review}, 58:101, 2025.

\bibitem{Ren:FasterRCNN:2017}
S.~{Ren}, K.~{He}, R.~{Girshick}, and J.~{Sun}.
\newblock {Faster R-CNN}: {T}owards real-time object detection with region proposal networks.
\newblock {\em PAMI}, 39(6):1137--1149, 2017.

\bibitem{Lin:Focal:2017}
Tsung-Yi Lin, Priya Goyal, Ross Girshick, Kaiming He, and Piotr Dollár.
\newblock Focal loss for dense object detection.
\newblock In {\em IEEE International Conference on Computer Vision}, pages 2999--3007, 2017.

\bibitem{khanam2024yolov11}
Rahima Khanam and Muhammad Hussain.
\newblock {Yolov11: An} overview of the key architectural enhancements.
\newblock {\em arXiv:2410.17725}, 2024.

\bibitem{Gilles:Detection:2018}
J.~Gilles, F.~Alvarez, N.~Ferrante, M.~Fortman, et~al.
\newblock Detection of moving objects through turbulent media. decomposition of oscillatory vs non-oscillatory spatio-temporal vector fields.
\newblock {\em Image and Vision Computing}, 73:40--55, 2018.

\bibitem{Zhang:Stabilization:2018}
Chao Zhang, Fugen Zhou, Bindang Xue, and Wenfang Xue.
\newblock Stabilization of atmospheric turbulence-distorted video containing moving objects using the monogenic signal.
\newblock {\em Signal Processing: Image Communication}, 63:19--29, 2018.

\bibitem{Lau:ATFaceGAN:2020}
Chun~Pong Lau, Hossein Souri, and Rama Chellappa.
\newblock {ATFaceGAN: S}ingle face image restoration and recognition from atmospheric turbulence.
\newblock In {\em FG}, pages 32--39, 2020.

\bibitem{Hu:object:2023}
Disen Hu and Nantheera Anantrasirichai.
\newblock Object recognition in atmospheric turbulence scenes.
\newblock In {\em 2023 31st European Signal Processing Conference (EUSIPCO)}, pages 561--565, 2023.

\bibitem{Bochkovskiy:YOLOv4:2020}
Alexey Bochkovskiy, Chien-Yao Wang, and Hong-Yuan~Mark Liao.
\newblock {YOLOv4: Optimal} speed and accuracy of object detection.
\newblock {\em ArXiv}, abs/2004.10934, 2020.

\bibitem{Uzun:Augmentation:2022}
Engin Uzun, Ahmet~An{\i}l Dursun, and Erdem Akag\"und\"uz.
\newblock Augmentation of atmospheric turbulence effects on thermal adapted object detection models.
\newblock In {\em CVPRW}, pages 241--248, June 2022.

\bibitem{Wang:YOLOR:2021}
Chien-Yao Wang, I-Hau Yeh, and Hong-Yuan~Mark Liao.
\newblock You only learn one representation: Unified network for multiple tasks.
\newblock {\em arXiv preprint arXiv:2105.04206}, 2021.

\bibitem{Zhang:VarifocalNet:2021}
Haoyang Zhang, Ying Wang, Feras Dayoub, and Niko Sünderhauf.
\newblock {VarifocalNet: An IoU}-aware dense object detector.
\newblock In {\em CVPR}, pages 8510--8519, 2021.

\bibitem{Feng:TOOD:2021}
Chengjian Feng, Yujie Zhong, Yu~Gao, Matthew~R. Scott, and Weilin Huang.
\newblock {TOOD: Task-aligned One-stage Object Detection}.
\newblock In {\em ICCV}, pages 3490--3499, 2021.

\bibitem{liu2021swin}
Ze~Liu, Yutong Lin, Yue Cao, Han Hu, Yixuan Wei, Zheng Zhang, Stephen Lin, and Baining Guo.
\newblock {Swin transformer: Hierarchical} vision transformer using shifted windows.
\newblock In {\em ICCV}, pages 10012--10022, 2021.

\bibitem{Gu:mamba:2023}
Albert Gu and Tri Dao.
\newblock {Mamba: Linear}-time sequence modeling with selective state spaces.
\newblock {\em arXiv preprint arXiv:2312.00752}, 2023.

\bibitem{lin2014microsoft}
Tsung-Yi Lin, Michael Maire, Serge Belongie, James Hays, et~al.
\newblock {Microsoft COCO: Common} objects in context.
\newblock In {\em ECCV}, pages 740--755, 2014.

\end{thebibliography}

\end{document}